\documentclass[letterpaper]{article} 
\usepackage{aaai2026}  
\usepackage{times}  
\usepackage{helvet}  
\usepackage{courier}  
\usepackage[hyphens]{url}  
\usepackage{graphicx} 
\usepackage{natbib}  
\usepackage{caption} 
\usepackage{algorithm}
\usepackage{algorithmic}
\usepackage{xcolor}
\usepackage{multirow}
\usepackage{caption}
\usepackage{amsmath}
\usepackage{cleveref}
\usepackage{amssymb}
\usepackage{booktabs}

\usepackage{newfloat}
\usepackage{listings}
\DeclareCaptionStyle{ruled}{labelfont=normalfont,labelsep=colon,strut=off} 
\floatstyle{ruled}
\newfloat{listing}{tb}{lst}{}
\floatname{listing}{Listing}
\title{MuSASplat: Efficient Sparse-View 3D Gaussian Splats via Lightweight Multi-Scale Adaptation}
\author{
    Muyu Xu\textsuperscript{\rm 1}, Fangneng Zhan\textsuperscript{\rm 2}, Xiaoqin Zhang\textsuperscript{\rm 3}, Ling Shao\textsuperscript{\rm 4}, Shijian Lu\textsuperscript{\rm 1}\\
}
\affiliations{
    \textsuperscript{\rm 1}Nanyang Technological University, Singapore\\
    \textsuperscript{\rm 2}Harvard University, USA\\
    \textsuperscript{\rm 3}Zhejiang University of Technology, China\\
    \textsuperscript{\rm 4}UCAS-Terminus AI Lab, University of Chinese Academy of Sciences, China\\
    muyu001@e.ntu.edu.sg, fnzhan@seas.harvard.edu, zhangxiaoqinnan@gmail.com\\ ling.shao@ieee.org, Shijian.Lu@ntu.ntu.edu.sg
}

\begin{document}

\maketitle

\begin{abstract}
Sparse-view 3D Gaussian splatting seeks to render high-quality novel views of 3D scenes from a limited set of input images. While recent pose-free feed-forward methods leveraging pre-trained 3D priors have achieved impressive results, most of them rely on full fine-tuning of large Vision Transformer (ViT) backbones and incur substantial GPU costs.
In this work, we introduce MuSASplat, a novel framework that dramatically reduces the computational burden of training pose-free feed-forward 3D Gaussian splats models with little compromise of rendering quality. Central to our approach is a lightweight Multi-Scale Adapter that enables efficient fine-tuning of ViT-based architectures with only a small fraction of training parameters. This design avoids the prohibitive GPU overhead associated with previous full-model adaptation techniques while maintaining high fidelity in novel view synthesis, even with very sparse input views.
In addition, we introduce a Feature Fusion Aggregator that integrates features across input views effectively and efficiently. Unlike widely adopted memory banks, the Feature Fusion Aggregator ensures consistent geometric integration across input views and meanwhile mitigates the memory usage, training complexity, and computational costs significantly.
Extensive experiments across diverse datasets show that MuSASplat achieves state-of-the-art rendering quality but has significantly reduced parameters and training resource requirements as compared with existing methods.
\end{abstract}

\section{Introduction}
\label{sec:intro}

\begin{figure}[t]
    \centering
    \includegraphics[width=0.48\textwidth]{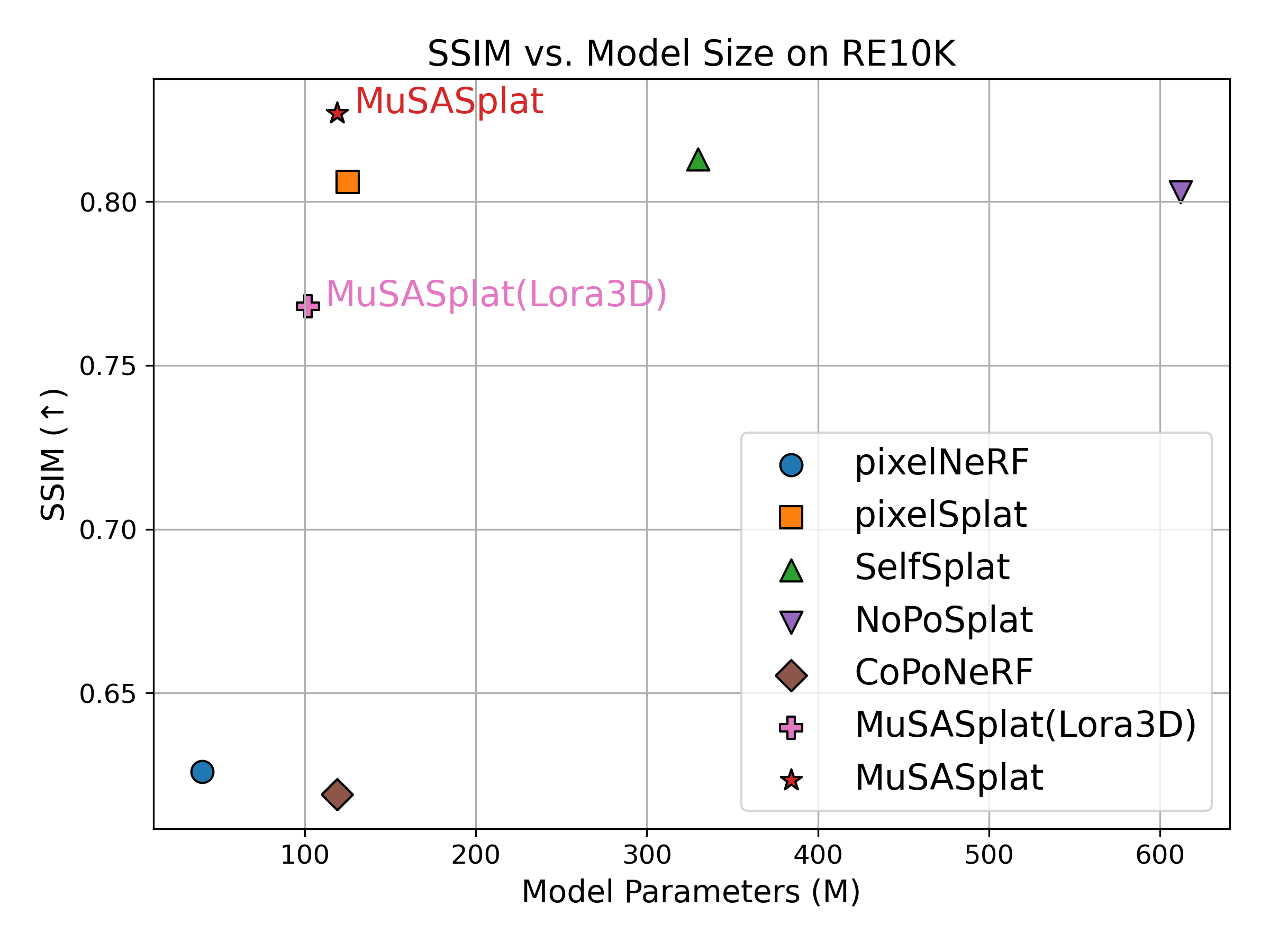}
    \caption{Unlike state-of-the-art sparse-view 3D rendering approaches that require either camera poses (like pixelSplat) or computationally intensive full model fine-tuning (like NoPoSplat), the proposed MuSASplat achieves superior reconstruction performance with unposed images and much reduced network parameters and computation costs. The experiments evaluate SSIM versus model size (in millions of parameters) on the RE10K dataset. We also report MuSASplat (LoRA3D), a variant where our adapter is replaced with LoRA3D~\cite{lu2024lora3d}, showing that the proposed Multi-Scale Adapter provides a clear advantage in accuracy.}
    \label{fig:teaser}
\end{figure}

Though 3D reconstruction with neural radiance field (NeRF) \cite{mildenhall2021nerf} and 3D Gaussian Splatting (3DGS) \cite{kerbl20233d} has achieved impressive novel-view synthesis, most existing methods still rely heavily on hundreds of posed images and per-scene optimization. The recent feed-forward networks~\cite{charatan2024pixelsplat, chen2024mvsplat, chen2024mvsplat360} obviate per-scene optimization but still require known camera poses, usually estimated by structure-from-motion algorithms such as COLMAP \cite{schonberger2016structure} which is computationally expensive and ill-suited under sparse-view settings. It remains a grand challenge to achieve robust and high-quality 3D scene reconstruction under the conditions of 1) few-shot unposed images, 2) without per-scene optimization, and 3) without high demand of GPU resources. 

Recent pose-free methods \cite{smart2024splatt3r, ye2024no, hong2024pf3plat, chen2024zerogs} leverage pretrained feed-forward 3D geometry networks \cite{zhang2025advances, wang2024dust3r, leroy2024grounding} to overcome the reliance on camera poses and extract point clouds directly from sparse views. For example, NoPoSplat \cite{ye2024no} combines DUSt3R~\cite{wang2024dust3r} with a 3D Gaussian head and achieves competitive performance through full-model fine-tuning. However, the full-model fine-tuning demands substantial GPU resources due to the need for adapting large vision transformer (ViT) backbones.

This paper presents \textit{MuSASplat}, a lightweight and scalable framework for pose-free sparse-view 3D reconstruction. MuSASplat explicitly targets the computational bottlenecks of existing methods, significantly reducing both training time and GPU memory usage while maintaining high-quality rendering, as shown in Figure \ref{fig:teaser}. It comes with two novel designs. First, it introduces a \textit{Multi-Scale Adapter} that enables efficient fine-tuning of ViT-based backbones. Unlike prior fine-tuning techniques such as LoRA~\cite{hu2022lora} that apply low-rank adaptations to linear projections, our adapter leverages the spatial structure implicit in ViT token sequences. Specifically, we rebuild the patchified token sequence back into a spatial feature map, preserving the image’s spatial layout, and apply a set of multi-scale depth-wise convolutions. These spatial-aware operations empower the adapter to better capture local geometric context across different receptive fields. By inserting a small number of lightweight convolutional layers into the transformer pipeline, this design achieves expressive adaptation capacity with minimal parameter overhead and significantly lowers GPU consumption compared to full-model fine-tuning.

Second, we introduce a \textit{Feature Fusion Aggregator} to replace conventional memory bank structures as used in mainstream multi-view 3D reconstruction. Existing memory bank designs, such as those in Spann3R~\cite{wang20243d} and CUT3R~\cite{wang2025continuous}, typically require sequentially encoding and decoding image pairs while updating a shared memory. Such processes are inefficient and memory-intensive, especially while handling dense input views. In contrast, our aggregator performs \textit{batch-wise} encoding and decoding of all input views simultaneously. It inserts a single lightweight aggregation module between the encoder and decoder to fuse features across viewpoints, enabling efficient global geometric consistency without iterative memory updates. This not only improves training throughput but also dramatically reduces GPU memory usage.
Experiments on multiple datasets show that MuSASplat achieves state-of-the-art performance in pose-free sparse-view 3D reconstruction, while requiring only a fraction of the training cost of existing methods.

The contributions of this work can be summarized in three major aspects:
\begin{itemize}
    \item We propose \textit{MuSASplat}, a lightweight framework for pose-free sparse-view 3D Gaussian splats that achieves strong rendering performance with minimal GPU overhead.
    \item We design a \textit{Multi-Scale Adapter} for ViT fine-tuning, which incorporates spatial-aware multi-scale depth-wise convolutions to improve scene understanding with a minimal number of parameters.
    \item We propose a \textit{Feature Fusion Aggregator} that enables efficient, batch-wise multi-view feature fusion without the need for iterative memory updates, improving both training efficiency and memory usage.
\end{itemize}

\section{Related Work}
\label{sec:related_work}

\subsection{2.1 Generalizable Novel View Synthesis}
NeRF\,\cite{mildenhall2021nerf} inaugurated neural rendering by modelling scenes as continuous radiance fields, but its reliance on hundreds of posed images and per-scene optimisation limits practical use.  3D Gaussian Splatting (3DGS)\,\cite{kerbl20233d} accelerates rendering by analytically rasterising anisotropic Gaussians, yet it inherits the same dense-capture requirement.  To alleviate per-scene training, a growing body of feed-forward methods~\cite{yu2021pixelnerf,wang2021ibrnet,chen2021mvsnerf,johari2022geonerf,xu2023wavenerf,charatan2024pixelsplat,chen2024mvsplat} build cost volumes or exploit epipolar geometry, but they still assume accurate camera poses and appreciable viewpoint overlap.  
The introduction of DUSt3R\,\cite{wang2024dust3r} moves a step further by predicting calibrated point maps directly from unposed stereo pairs.  Built on this backbone, Splatt3R\,\cite{smart2024splatt3r} attaches a Gaussian decoder yet updates only a handful of weights, whereas NoPoSplat\,\cite{ye2024no} fully fine-tunes the network and attains the current best fidelity at the cost of heavy GPU usage.  The present work continues the pose-free direction but replaces full-model tuning with an efficient adapter and aggregator, thereby slashing training cost while preserving high quality.

\subsection{2.2 Parameter-Efficient Fine-Tuning}
Low-rank adaptation (LoRA)\,\cite{hu2022lora} demonstrates that inserting rank-constrained matrices into otherwise frozen Transformers transfers knowledge with only a few per-cent of the original parameters; QR-LoRA\,\cite{yang2025qr} further stabilises the optimiser by factorising the LoRA weights using QR decomposition.  More recently, the “5\% \textgreater 100\%” study\,\cite{yin20255} shows that multi-scale depth-wise adapters can rival or even surpass full fine-tuning on vision tasks while adding roughly five per-cent parameters, and Adapter-X\,\cite{li2024adapter} extends the idea to 3D point clouds.  In the 3D domain specifically, LoRA3D\,\cite{lu2024lora3d} affirms that parameter-efficient techniques can adapt neural rendering pipelines with limited hardware.  Guided by these findings, our Multi-Scale Adapter rebuilds ViT tokens back into feature maps and applies multi-scale depth-wise convolutions, injecting spatial priors at a minimal parameter cost.

\subsection{2.3 Efficient Multi-View Feature Aggregation}
Consistent fusion of unposed views is commonly addressed with external memory structures.  Spann3R\,\cite{wang20243d} and CUT3R\,\cite{wang2025continuous} sequentially process image pairs and update a global memory bank, which leads to $V-1$ times encoder–decoder passes and high memory footprints.  PreF3R\,\cite{chen2024pref3r} incrementally merges variable-length sequences but still performs iterative updates.  In contrast, our Feature Fusion Aggregator encodes all views in a single batch and inserts just one lightweight fusion layer, so aggregation latency remains constant and GPU memory usage stays low even when the number of input images grows.

\section{Method}
\label{sec:method}

\begin{figure*}[t]
    \centering
    \includegraphics[width=\textwidth]{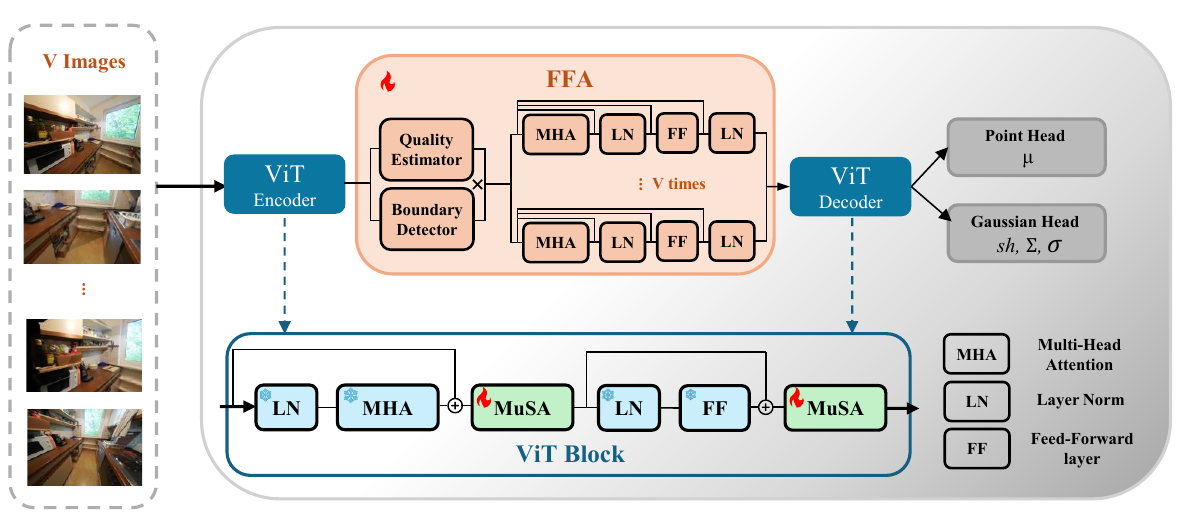}
    \caption{Overview of the MuSASplat architecture. Given unposed multi-view input images, we extract per-view features using a ViT encoder. ViT consists of multiple stacked ViT blocks, which are the basic computational units containing self-attention and feed-forward sublayers. Our proposed Multi-Scale Adapter (MuSA) modules are inserted into the blocks to enhance spatial awareness while introducing minimal extra parameters. The resulting features from different views are fused in a single forward pass by the Feature Fusion Aggregator (FFA), which adaptively integrates geometric information using view-specific quality estimator and boundary detector as elaborated in section 3.4. The fused feature is then decoded by a lightweight ViT decoder and passed to a point head and a Gaussian head to generate the parameters of 3D Gaussian primitives for rendering.}
    \label{fig:pipeline}
\end{figure*}
Let $\mathcal{I}=\{I^{v}\}_{v=1}^{V}$ be $V$ unposed RGB images.  
Our MuSASplat predicts a set of anisotropic Gaussians
\begin{equation}  
    \mathcal{G}=\{\mathrm{G}_{n}\}_{n=1}^{N},
    \;\mathrm{G}_{n}=(\boldsymbol\mu_{n},\alpha_{n},
      \boldsymbol\Sigma_{n},\mathbf{sh}_{n}) ,
\end{equation}

\noindent which are splatted for novel-view synthesis.  
Only three lightweight modules, the \emph{Multi-Scale Adapter (MuSA)}, the \emph{Feature Fusion Aggregator (FFA)}, and the Gaussian/point heads, are trainable. The large ViT backbone remains frozen.  
Figure~\ref{fig:pipeline} sketches the whole pipeline.
\subsection{3.1 Preliminaries}
\label{subsec:preliminary}
\textbf{3D Gaussian Splatting (3D-GS).} 
3DGS~\cite{kerbl20233d} represents a scene's radiance field using a set of anisotropic 3D Gaussians, each encoding the radiance of its surrounding region. Each Gaussian is parameterized by its mean position $\boldsymbol{\mu} \in \mathbb{R}^3$, opacity $\alpha \in \mathbb{R}$, covariance $\boldsymbol{\Sigma} \in \mathbb{R}^{3 \times 3}$, and SH $\boldsymbol{sh} \in \mathbb{R}^k$ representing color. 
However, traditional 3D-GS fits Gaussian splats to a scene through iterative optimization and is unsuitable for generalizable feed-forward models.

Recent generalizable 3D-GS methods~\cite{charatan2024pixelsplat, chen2024mvsplat, szymanowicz2024flash3d, szymanowicz2024splatter} directly predict pixel-aligned 3D Gaussians from a set of $N$ images $I=\{I_i\}_{i=1}^{N}$. 
However, these approaches rely on known camera parameters, which limits their applicability to calibrated settings. In contrast, our method estimates per-view point maps and projects them to a canonical space, enabling consistent 3D Gaussian supervision directly from uncalibrated RGB images.

\noindent\textbf{DUSt3R.}  DUSt3R~\cite{wang2024dust3r} is a ViT-based approach that jointly solves camera calibration and 3D reconstruction using only images. Given two input images, it predicts dense 3D point clouds, referred to as \textit{pointmaps}, which establish a per-pixel 2D-to-3D mapping. Specifically, a pointmap $X_{a,b}$ maps each pixel $i = (u, v)$ in image $I_a$ to a corresponding 3D point $X_{a,b}^{u,v}$, expressed in the coordinate system of camera $C_b$. By regressing two pointmaps $X_{a,a}$ and $X_{b,a}$ in a shared coordinate frame ($C_a$), DUSt3R effectively performs joint calibration and 3D reconstruction.

\subsection{3.2 Multi-Scale Adapter (MuSA)}
\begin{figure}[h]
    \centering
    \includegraphics[width=0.48\textwidth]{figs/musa_layer_v4.pdf}
    \caption{
    Comparison between LoRA and our proposed MuSA layer.  
    \textbf{Top}: LoRA injects a low-rank residual update into the frozen pre-trained model via two linear layers, operating purely in the token space without spatial awareness.  
    \textbf{Bottom}: MuSA reconstructs the spatial layout of tokens into a feature map, applies depth-wise convolutions at multiple kernel sizes to capture local structure, and projects the adapted features back into the token stream. This design enables spatial reasoning while maintaining parameter efficiency.
    }
    \label{fig:musa_layer}
\end{figure}
Low-Rank Adaptation (LoRA)~\cite{hu2022lora} is a widely adopted strategy for parameter-efficient fine-tuning. As illustrated in Figure \ref{fig:musa_layer}, LoRA modifies a frozen model by injecting two lightweight linear layers into the residual path, effectively learning a low-rank update to the output of a linear projection. This approach works well in language modeling and some 2D vision tasks where token-wise dependencies dominate. However, its effectiveness is limited in 3D vision tasks such as novel view synthesis, where spatial coherence and geometric consistency are essential.

The core limitation lies in LoRA's inability to reintroduce spatial priors. Since a vanilla ViT processes each image as a sequence of independent patch tokens, it discards the 2D structure within and between patches. LoRA, operating only on linear transformations of these 1D tokens, does not recover any notion of proximity, local structure, or continuity across the image plane. Consequently, when applied to sparse-view 3D reconstruction, LoRA-equipped models often generate inconsistent geometry, overfit to local appearance cues, or hallucinate non-existent structures.

To overcome these limitations, we propose MuSA: a Multi-Scale Adapter that restores spatial reasoning to the ViT encoding process, while maintaining parameter efficiency. The key idea is to reinterpret the token sequence as a feature map and process it using depth-wise convolutions with multiple kernel sizes. In this way, MuSA enables the model to reason about local and global spatial relationships, which LoRA cannot capture natively.

Concretely, for each input image $I^v$, the frozen encoder produces a token sequence $\mathbf{X}^{v} \in \mathbb{R}^{N \times C}$, where $N = HW / P^2$ is the number of patches and $C$ the channel dimension. We first project the tokens to a reduced dimension via a $1\times1$ linear projection:  
\begin{equation}
\mathbf{Y}^{v} = \mathrm{reshape}(\mathbf{X}^{v} \mathbf{W}_{\!\downarrow}) \in \mathbb{R}^{C' \times h \times w},
\end{equation}
where $C' = C / r$ with reduction ratio $r=4$, and $h \times w = N$ corresponds to the spatial grid of patches.  

We then apply three depth-wise convolutions with kernel sizes $3$, $5$, and $7$, and average their outputs to obtain:
\begin{equation}
\mathbf{Z}^{v} = \tfrac{1}{3} \sum_{k \in \{3, 5, 7\}} \mathrm{DWConv}_{k}(\mathbf{Y}^{v}).
\end{equation}
This multi-scale aggregation captures both fine and coarse spatial dependencies within the patch grid. The result is passed through a point-wise convolution and a GELU activation, then projected back to the original dimension:
\begin{equation}
\widehat{\mathbf{X}}^{v} = \left(\mathrm{GELU}\left(\mathrm{PWConv}(\mathbf{Z}^{v}) \right) \right) \mathbf{W}_{\!\uparrow}.
\end{equation}
We finally add the residual to the frozen token stream:
\begin{equation}
\mathbf{X}^{v}_{\text{out}} = \mathbf{X}^{v} + \widehat{\mathbf{X}}^{v}.
\end{equation}
All learnable parameters are zero-initialized so that the model starts from the behavior of the pre-trained backbone, ensuring stability at the start of fine-tuning.

\subsection{3.3 Mini-Grid Branch in MuSA}

One might suspect that further gains could come from recovering \emph{intra-patch} structure, whose information is irretrievably lost once a $16\times16$ region is collapsed into a single token.
We therefore insert a \textit{mini-grid branch} that maps each token into a small $p\times p$ feature map (here $p{=}4$), applies one depth-wise convolution, and averages the result back into a token-sized residual.
Because the original spatial arrangement of pixels inside a patch is unknown, the branch effectively has to \emph{relearn} a permutation mapping from scratch, a task that is under-constrained by sparse 3D supervision.
In practice, its weights remain near-zero and the branch yields no measurable improvement on any dataset we tested.
The negative result is nevertheless informative: it reveals that the primary benefit of our adapter is not recovering sub-patch detail but rather exploiting the cross-patch spatial relations that ViT discards.
In other words, the small, multi-scale convolutions are sufficient to inject the geometric cues that the 3D Gaussian decoder needs.

\subsection{3.4 Feature Fusion Aggregator (FFA)}
\label{sec:ffa}

Some existing multi-view pose-free pipelines, such as Spann3R~\cite{wang20243d} and CUT3R~\cite{wang2025continuous}, rely on an external memory bank that is updated once per image pair.
The memory must be read and written \(O(V)\) times, which slows training when the number of input views \(V>2\) and stores features from every view, inflating GPU memory.

Instead of iterative memory updates, we process \emph{all} input views simultaneously in each batch.
The encoded features from all views are stacked as
    $\mathbf{F} \in \mathbb{R}^{B \times V \times L \times C}$,
where $V$ is the number of views.
To assess the per-token quality, a lightweight MLP (\textit{quality estimator}) estimates a confidence score for each token:
\begin{equation}
    q^{v}_{\ell} = \sigma(\mathbf{w}_q^\top \mathbf{f}^{v}_{\ell}).
\end{equation}
In parallel, a second MLP (\textit{boundary detector}) detects boundary-view indicators from global pooled features $\bar{\mathbf{f}}^v = \frac{1}{L} \sum_{\ell} \mathbf{f}^v_\ell$, producing a binary mask $b^v \in \{0,1\}$ to identify potentially valuable views. A tunable weight $\lambda > 1$ boosts their contribution.
The final attention weights are modulated by both token-wise confidence and boundary-view weight:
\begin{equation}
    w^{v}_{\ell} = \begin{cases}
      q^v_\ell \cdot \lambda & \text{if } b^v = 1 \\
      q^v_\ell & \text{otherwise}
    \end{cases}.
\end{equation}
Low-confidence tokens ($q^v_\ell < \tau$) are masked. For each query view $v$, the remaining views serve as key/value pairs:
\begin{equation}
    \mathbf{A}^{v}= \mathrm{softmax}\!
      \left( \frac{\mathbf{Q}^{v}(\mathbf{K}^{v})^\top}{\sqrt{d}} 
             + \log \mathbf{M}^{v} \right),\quad
    \widetilde{\mathbf{F}}^{v}= \mathbf{A}^{v}\mathbf{V}^{v}.
\end{equation}
Finally, we fuse the attended features with the original using an MLP residual connection:
\begin{equation}
    \widehat{\mathbf{F}}^{v} = \mathbf{F}^{v} 
    + \mathrm{MLP}\left( \left[ \mathbf{F}^{v}, \widetilde{\mathbf{F}}^{v} \right] \right).
\end{equation}

Because the aggregator is invoked only once, its latency is constant with respect to \(V\), and its memory footprint is bounded by the batch itself.
Replacing the memory bank with this lightweight aggregator reduces peak GPU memory by 0.3$\times$ and accelerates training by 4.2$\times$.

\subsection{3.5 Point-Cloud Viewpoint Augmentation}
\label{sec:pca}
When the input has only two views ($V{=}2$), the Gaussian count $N\!\propto\!V$ is small, which can result in uncovered regions appearing as holes or over-large Gaussians if the input views have small overlap (<30\%).
To mitigate this issue, we predict an initial sparse point cloud $\mathcal{P}$ and relative pose $\{\mathbf{T}_{v}\}$ from the \textbf{same frozen ViT backbone followed by a point head}
\begin{equation}
    \bigl[\mathcal{P},\,\{\mathbf{T}_{v}\}\bigr]
      = g_{\boldsymbol\phi}\bigl( I^{1}, I^{2}\bigr),
\end{equation}
where $g_{\boldsymbol\phi}$ is fixed during Gaussian training.
Each 3-D point $\mathbf{p}\!\in\!\mathcal{P}$ is rasterised from $K$ interpolated camera poses $\{\mathbf{T}_{k}\}$ to form synthetic images
$
    \widetilde{I}^{k} = R(\mathcal{P},\mathbf{T}_{k}).
$

The synthetic views are appended to $\mathcal{I}$, so the total Gaussian count increases and our model can learn to fill the holes with the extra Gaussians.  
During training we add a multi-view consistency term
\begin{equation}
    \mathcal{L}_{\text{aug}} 
      = \tfrac{1}{K}\sum_{k=1}^{K}
        \bigl\|
          \Pi\bigl(\mathcal{G};\mathbf{T}_{k}\bigr) - \widetilde{I}^{k}
        \bigr\|_{1},
\end{equation}
where $\Pi$ denotes differentiable Gaussian splatting.  
The full objective is
\begin{equation}
    \mathcal{L}
      = \lambda_{\text{rgb}}\mathcal{L}_{\text{rgb}}
      + \lambda_{\text{aug}}\mathcal{L}_{\text{aug}},
\end{equation}
with $\lambda_{\text{aug}}\!=\!0.05$. The RGB loss $\mathcal{L}_{\text{rgb}}$ follows NoPoSplat and consists of a weighted sum of mean squared error (MSE) and perceptual LPIPS loss ($\lambda_{\text{mse}}=1.0, \lambda_{\text{lpips}}=0.2$) between rendered and ground-truth images.

\section{Experiments}
\label{sec:experiment}
\begin{table*}[t]
\centering
\renewcommand{\arraystretch}{1.2}
\setlength{\tabcolsep}{0.9mm}
\begin{tabular}{ll|c|ccc|ccc}
\toprule
& \textbf{Method} 
& Params(M)
& \multicolumn{3}{c|}{RE10K} 
& \multicolumn{3}{c}{ACID} \\
& && PSNR$\uparrow$ & SSIM$\uparrow$ & LPIPS$\downarrow$
  & PSNR$\uparrow$ & SSIM$\uparrow$ & LPIPS$\downarrow$ \\
\midrule
\multicolumn{8}{l}{\textit{2-View Reconstruction}} \\
\midrule
\multirow{4}{*}{Pose-Required}
& pixelNeRF~\cite{yu2021pixelnerf}  & -         & 19.824 & 0.626 & 0.485 & 20.323 & 0.561 & 0.533 \\
& AttnRend~\cite{du2023learning}   & -         & 22.664 & 0.762 & 0.269 & 24.475 & 0.730 & 0.287 \\
& pixelSplat~\cite{charatan2024pixelsplat}   & -       & 23.848 & 0.806 & 0.185 & \textit{25.819} & 0.779 & \underline{0.195} \\
& MVSplat~\cite{chen2024mvsplat} & -            & 23.977 & \textit{0.811} & \underline{0.176} & 25.512 & 0.773 & \textit{0.196} \\
\midrule
\multirow{7}{*}{Pose-Free}
& DUSt3R~\cite{wang2024dust3r}    & -          & 15.382 & 0.447 & 0.432 & 16.286 & 0.411 & 0.447 \\
& MASt3R~\cite{leroy2024grounding}  & -           & 14.907 & 0.431 & 0.452 & 16.179 & 0.409 & 0.461 \\
& Splatt3R~\cite{smart2024splatt3r}  & 80    & 15.318 & 0.490 & 0.425 & 16.754 & 0.472 & 0.448 \\
& CoPoNeRF ~\cite{hong2024unifying}& 286     & 18.938 & 0.619 & 0.226 & 20.950 & 0.606 & 0.406 \\
& SelfSplat ~\cite{kang2025selfsplat}  & 330  & \textit{24.220} & \underline{0.813} & 0.188 & \textbf{26.710} & \textbf{0.801} & \textit{0.196} \\
& NoPoSplat~\cite{ye2024no} &612   & \textbf{24.833} & 0.803 & \textbf{0.172} & \underline{25.961} & \underline{0.781} & 0.205 \\
& MuSASplat (Ours) &119 & \underline{24.324} & \textbf{0.827} & \textit{0.182} & 25.802 & \textit{0.780} & \textbf{0.189} \\
\midrule
\multicolumn{8}{l}{\textit{5-View Reconstruction}} \\
\midrule
\multirow{3}{*}{Pose-Free}
& CUT3R + GS Head  ~\cite{wang2025continuous}  &170     & \textit{21.751} & \textit{0.701} & \underline{0.284} & \textit{22.981} & \textit{0.782} & \textit{0.223} \\
& PREF3R ~\cite{chen2024pref3r} & 161                & \underline{22.181} & \underline{0.749} & \textit{0.298} & \underline{23.212} & \underline{0.796} & \underline{0.191} \\
& MuSASplat (Ours)&119  & \textbf{24.721} & \textbf{0.832} & \textbf{0.155} & \textbf{26.362} & \textbf{0.812} & \textbf{0.171} \\
\bottomrule
\end{tabular}
\caption{Quantitative comparison on RE10K and ACID datasets under 2-view and 5-view settings. \textbf{The best results are formatted in bold, the second-best results are underlined, and the third-best results are italicized.} MuSASplat has a competitive performance with both pose-required and pose-free baselines in 2-view settings, and demonstrates superior performance in 5-view comparisons against memory-based methods like PREF3R and CUT3R. To ensure fair comparison of model efficiency, the parameter count column is only reported for pose-free methods. DUSt3R and MASt3R are pre-trained on external datasets for point cloud prediction and are included solely for inference-time comparison, so their parameter counts are not listed. For the 5-view setting, we freeze the backbone of each model and retrain only the remaining modules to better isolate and compare the efficiency and performance of memory bank versus feature fusion aggregator designs.}
\label{tab:main_table}
\end{table*}
We validate the effectiveness of our proposed MuSASplat through extensive experiments. We first present our training setup and datasets, followed by main comparisons against prior methods, and finally evaluate the contribution of each component via ablations and targeted substitution studies.

\subsection{4.1 Training and Evaluation Setup}
\label{subsec:exp_setup}

\noindent\textbf{Datasets.}  
To ensure consistency with prior work, we adopt the same training and evaluation protocol as NoPoSplat~\cite{ye2024no}. We train on three large-scale pose-free datasets: RealEstate10K (RE10K)~\cite{zhou2018stereo} and ACID~\cite{liu2021infinite}, which collectively span indoor and outdoor scenes across diverse conditions. Evaluation on RE10K and ACID follows NoPoSplat's overlap-based partitioning.

\noindent \textbf{Baselines.}
We compare against a range of generalizable 3D reconstruction methods under both two-view and multi-view settings. For the two-view input case, we evaluate pose-required models such as PixelNeRF~\cite{yu2021pixelnerf}, AttnRend~\cite{du2023learning}, PixelSplat~\cite{charatan2024pixelsplat}, and MVSplat~\cite{chen2024mvsplat}, as well as recent pose-free methods including Splatt3R~\cite{smart2024splatt3r}, CoPoNeRF~\cite{hong2024unifying}, NoPoSplat~\cite{ye2024no}, and SelfSplat~\cite{kang2025selfsplat}. In addition, to further assess the effectiveness of our proposed feature fusion aggregator, we conduct experiments using five-view input and compare against methods that rely on traditional memory banks. Specifically, we include PREF3R~\cite{chen2024pref3r} and CUT3R~\cite{wang2025continuous} equipped with a Gaussian head. Since PREF3R is not open-sourced, we re-implement its architecture based on the original paper, using Spann3R~\cite{wang20243d} as the backbone and appending a Gaussian rendering head. The model is retrained under the same setting as ours for a fair comparison. These comparisons provide a comprehensive evaluation across both input regimes and highlight the generalizability of our approach. 

\noindent\textbf{MuSA Adapter Extension to Other Models.}  
To test the generalizability of our adapter design, we inject our proposed MuSA Adapter into the transformer blocks of both NoPoSplat and SelfSplat, resulting in NoPoSplat-MuSA and SelfSplat-MuSA variants. These are compared against LoRA3D~\cite{lu2024lora3d} inserted into the same backbones under identical settings. All adapter comparisons are conducted on RE10K.

\begin{figure*}[t]
    \centering
    \includegraphics[width=0.9\textwidth]{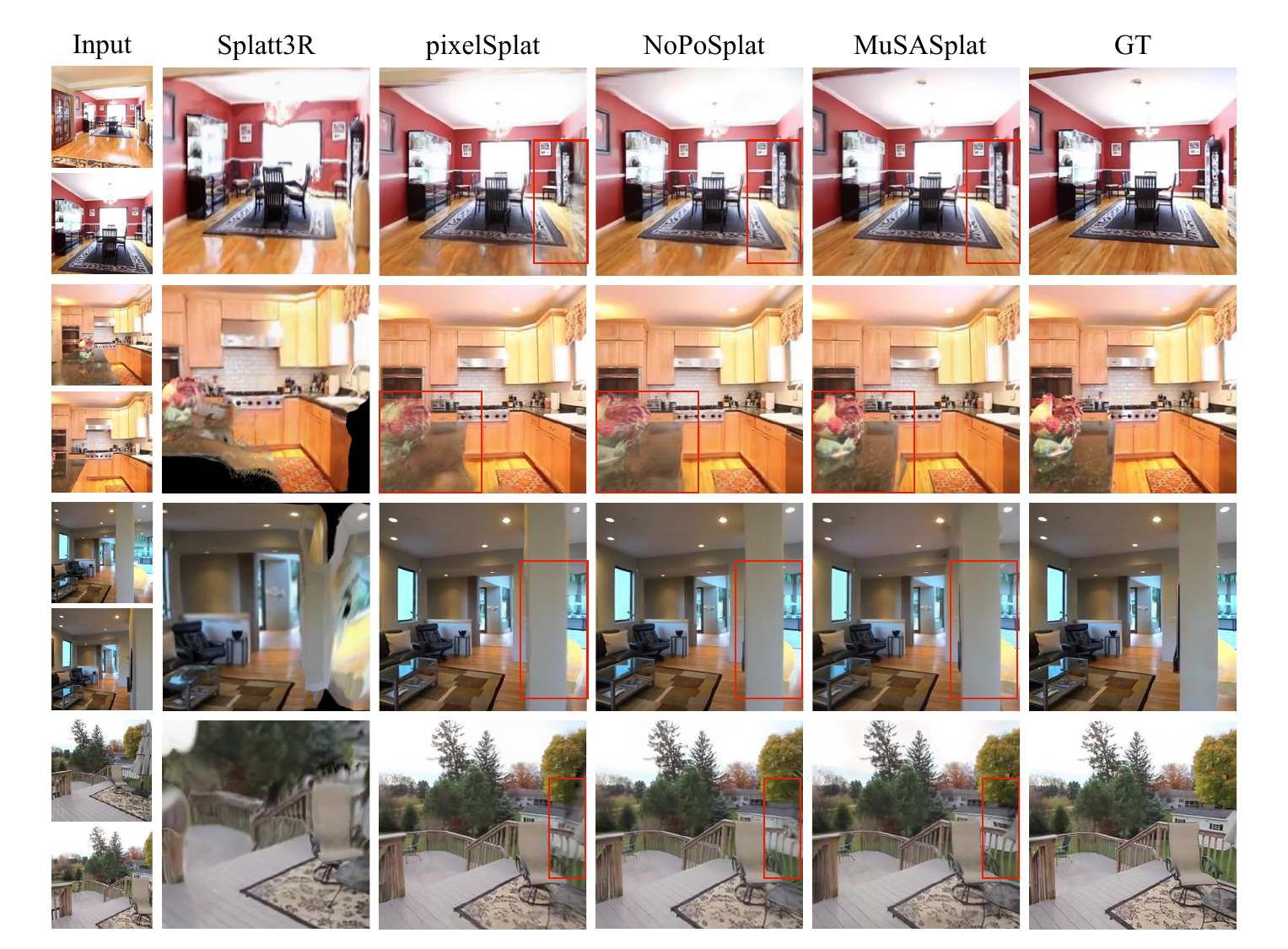}
    \caption{Qualitative comparison on RE10K. Compared to NoPoSplat and PixelSplat, our MuSASplat yields more complete geometry and fewer artifacts in occluded regions. The major differences are highlighted with red boxes.}
    \label{fig:qualitative_main}
\end{figure*}
\noindent\textbf{Implementation Details.}  
Our model is implemented in PyTorch and trained using AdamW with learning rate $5\cdot10^{-5}$, weight decay $0.05$, and gradient clipping of $0.5$.  
We freeze all pre-trained ViT backbone weights and only train the Multi-Scale Adapter, Feature Fusion Aggregator, and Gaussian/point heads.  
All images are resized to $256\times256$ resolution.  
Training takes place on a single RTX3090 GPU with a batch size of 4.  Training lasts 75K iterations.

\subsection{4.2 Main Results}
\label{subsec:exp_results}

\noindent\textbf{Quantitative Results.}  
We report PSNR, SSIM, and LPIPS~\cite{zhang2018unreasonable} on both the RE10K and ACID datasets in Table~\ref{tab:main_table}.
MuSASplat consistently outperforms all pose-required baselines across both datasets, achieving +0.35 dB higher PSNR than the best pose-required method (MVSplat) on RE10K and +0.29 dB on ACID.
Although NoPoSplat achieves slightly better PSNR on RE10K, our method delivers more balanced performance across all metrics, including a lower LPIPS on ACID (0.189 vs. 0.205), indicating more perceptually faithful reconstructions.
In the 5-view setting, MuSASplat surpasses both memory-based pose-free baselines by a significant margin, outperforming PREF3R by +2.5 dB PSNR and CUT3R by +3.0 dB on RE10K.
It also achieves the best SSIM and LPIPS on both datasets, showcasing the effectiveness of our feed-forward multi-view aggregation in high-visibility settings.

\noindent\textbf{Qualitative Results.}  
As shown in Fig. \ref{fig:qualitative_main}, MuSASplat recovers more complete geometry and precise textures in both indoor and outdoor scenes.  
Notably, fewer floaters and holes appear around edges and occlusion boundaries as highlighted with the red boxes.

\subsection{4.3 Component Analysis and Ablation Studies}
\label{subsec:abl_study}

\noindent\textbf{Ablation studies.}  
We conduct ablation studies to examine the contributions of each module in MuSASplat, including the Multi-Scale Adapter, the Feature Fusion Aggregator, and the point cloud augmentation strategy. We also construct a variant named MuSASplat (LoRA3D) by replacing our adapter with LoRA3D~\cite{hu2022lora} which uses LoRA to fine-tune large 3D geometric foundation models.
As shown in Table~\ref{tab:ablation}, removing the Multi-Scale Adapter leads to a significant performance drop of 4.2 dB. Using Lora3D instead of our Multi-Scale Adapter also leads to a drop of 1.4 dB, confirming the importance of spatial-aware feature adaptation.
Without the Aggregator, performance drops by 0.8 dB despite having similar parameter count and training speed, showing that single-pass feature fusion is more effective than independent view processing.

We also evaluate the impact of our point cloud (pcd) augmentation strategy by disabling it during training. This variant keeps the same model size and training speed, since the augmentation is performed as a pre-processing step. While the performance drop is minor (about 0.3 dB), this is expected because most scenes already have high input-view overlap. The augmentation mainly benefits cases with large viewpoint shifts, where it helps fill in holes that would otherwise remain unobserved.

Finally, another variant MuSASplat(Memory Bank) obtained by replacing the Aggregator with a memory bank mechanism similar to Spann3R causes a large reduction in training efficiency: the frame processing speed drops from 1.81 it/s (iteration per second) to 0.43 it/s, and the model size increases by over 30\%. However, this change does not yield any performance gain, highlighting the efficiency advantage of our feed-forward design.

\begin{table}[t]
    \centering
    \resizebox{0.46\textwidth}{!}{
    \begin{tabular}{l|c|c|c}
        \toprule
        Variant & PSNR↑ & Speed (it/s) & Params (M) \\
        \midrule
        MuSASplat (full) & 24.32 & 1.81 & 119 \\
        w/o Adapter       & 20.12 & 2.02 & 92 \\
        w/o Aggregator  & 23.50 & 1.96 & 113 \\
        w/o Pcd Augmentation & 24.03 & 1.81 & 119 \\
        MuSASplat(Lora3D) & 22.89 & 1.65 & 102 \\
        MuSASplat(Memory Bank) & 23.66 & 0.43 & 155 \\
        \bottomrule
    \end{tabular}
    }
    \caption{Ablation study on MuSA and FFA. Removing either module or using Lora3D instead of our MuSA layer hurts performance. Replacing our aggregator with the memory bank further drastically reduces training efficiency.}
    \label{tab:ablation}
\end{table}
\begin{table}[t]
    \centering
    \begin{tabular}{l|c|c|c}
        \toprule
        Model Variant & PSNR↑ & SSIM↑ & LPIPS↓ \\
        \midrule
        NoPoSplat-LoRA3D & 21.72 & 0.741 & 0.283 \\
        NoPoSplat-MuSA   & \textbf{23.39} & \textbf{0.794} & \textbf{0.221} \\
        \midrule
        SelfSplat-LoRA3D & 21.60 & 0.748 & 0.271 \\
        SelfSplat-MuSA   & \textbf{23.14} & \textbf{0.799} & \textbf{0.223} \\
        \bottomrule
    \end{tabular}
    \caption{MuSA Adapter vs. LoRA3D on RE10K. Replacing LoRA3D by MuSA Adapter consistently improves performance in both NoPoSplat and SelfSplat.}
    \label{tab:adapter_transfer}
\end{table}
\noindent\textbf{Adapter Transferability compared with Lora.} 
To assess the general applicability of the MuSA Adapter across different architectures, we compare it against LoRA3D by integrating both adapters into two distinct base models: NoPoSplat and SelfSplat. Specifically, we insert either MuSA or LoRA3D into the backbone of each model and freeze the pretrained weights of the backbones, training each model with the rest of the architecture unchanged. This allows for a fair evaluation of the fine-tuning mechanism itself. As shown in Table~\ref{tab:adapter_transfer}, replacing LoRA3D with MuSA consistently improves performance in terms of PSNR, SSIM, and LPIPS on both models. While SelfSplat employs a Swin Transformer backbone that preserves spatial structures, we adapt our MuSA design by omitting the reshaping step to match the architectural differences. Despite this, MuSA still yields notable gains, demonstrating its versatility and effectiveness across various encoder types.

\section{Conclusion}
\label{sec:conclusion}
We propose MuSASplat, a lightweight framework for pose-free sparse-view 3D Gaussian reconstruction that delivers competitive rendering quality with substantially lower GPU memory usage and training cost. Our Multi-Scale Adapter enables spatially aware fine-tuning of ViT backbones, while the Feature Fusion Aggregator efficiently fuses multi-view features in a single pass. Together, they achieve strong performance with 5× fewer parameters than existing pose-free baselines.
Despite its effectiveness, MuSASplat is currently tailored to ViT-based 3D models and may not fully exploit newer architectures such as VGGT~\cite{wang2025vggt}. Furthermore, it is designed for static scene reconstruction, and extending it to dynamic settings remains a promising direction.

\section{Acknowledgments}
This work is funded by the Ministry of Education, Singapore, under the Tier-1 project scheme with a project number RT18/22.

\bibliography{aaai2026}

\end{document}